\documentclass[10pt, conference]{IEEEtran}
\IEEEoverridecommandlockouts
\usepackage{cite}
\usepackage{amsmath,amssymb,amsfonts}
\usepackage[bookmarks=false]{hyperref}
\usepackage{graphicx}
\usepackage{textcomp}
\usepackage{xcolor}
\usepackage{bm}
\usepackage{dsfont}
\usepackage{comment}
\usepackage{algorithm}
\usepackage{algpseudocode}

\definecolor{darkspringgreen}{rgb}{0.09, 0.45, 0.27}
\def\BibTeX{{\rm B\kern-.05em{\sc i\kern-.025em b}\kern-.08em
    T\kern-.1667em\lower.7ex\hbox{E}\kern-.125emX}}

\makeatletter
 \let\old@ps@headings\ps@headings
 \let\old@ps@IEEEtitlepagestyle\ps@IEEEtitlepagestyle
 \def\confheader#1{%
 \def\ps@headings{%
 \old@ps@headings%
 \def\@oddhead{\strut\hfill#1\hfill\strut}%
 \def\@evenhead{\strut\hfill#1\hfill\strut}%
 }%
 \def\ps@IEEEtitlepagestyle{%
 \old@ps@IEEEtitlepagestyle%
 \def\@oddhead{\strut\hfill#1\hfill\strut}%
 \def\@evenhead{\strut\hfill#1\hfill\strut}%
 }%
 \ps@headings%
 }
 \makeatother

\confheader{Published as a conference paper at 2021 58th ACM/IEEE DAC(\url{https://doi.org/10.1109/DAC18074.2021.9586187})
}

\begin{document}
\title{Pruning In Time (PIT): A Lightweight\\Network Architecture Optimizer for\\Temporal Convolutional Networks
}


\author{\IEEEauthorblockN{Matteo Risso\IEEEauthorrefmark{1}, Alessio Burrello\IEEEauthorrefmark{2}, Daniele Jahier Pagliari\IEEEauthorrefmark{1}, Francesco Conti\IEEEauthorrefmark{2}, \\Lorenzo Lamberti\IEEEauthorrefmark{2},  Enrico Macii\IEEEauthorrefmark{1}, Luca Benini\IEEEauthorrefmark{2}, Massimo Poncino\IEEEauthorrefmark{1}}
\IEEEauthorblockA{\IEEEauthorrefmark{1}Politecnico di Torino, Turin, Italy. \textit{Email: name.surname@polito.it} \\
\IEEEauthorrefmark{2}University of Bologna, Bologna, Italy. \textit{Email: name.surname@unibo.it} 
}}

\maketitle
\begin{abstract}
Temporal Convolutional Networks (TCNs) are promising Deep Learning models for time-series processing tasks. One key feature of TCNs is time-dilated convolution, whose optimization requires extensive experimentation. We propose an automatic dilation optimizer, which tackles the problem as a weight pruning on the time-axis, and learns dilation factors together with weights, in a single training. Our method reduces the model size and inference latency on a real SoC hardware target by up to 7.4$\times$ and 3$\times$, respectively with no accuracy drop compared to a network without dilation. It also yields a rich set of Pareto-optimal TCNs starting from a single model, outperforming hand-designed solutions in both size and accuracy. 
\end{abstract}

\section{Introduction}

Deep learning (DL) models achieve state-of-the-art performance in many time-series processing tasks, such as bio-signals analysis~\cite{temponet_2019, epilepsy_CNN_2018}, predictive maintenance~\cite{shm_dnn_review_2020} and sound classification~\cite{dnn_speakerdetection_2018}. Recurrent Neural Networks (RNNs) have long been considered the de facto standard for these tasks~\cite{rnn_review_2015}, but recently, Temporal Convolutional Networks (TCNs) -- a subclass of Convolutional Neural Networks (CNNs) dedicated to time series -- have been shown to provide comparable accuracy, whilst offering several advantages from a computational standpoint: smaller memory footprint, more data reuse opportunities and higher arithmetic intensity~\cite{tcn_original_2018}.

The deployment of DL models for direct inference on Internet of Things (IoT) edge devices is an emerging paradigm for the aforementioned time-series analysis tasks, as it provides several benefits compared to a cloud-centric approach. Edge computing reduces the pressure on the network, hence improving scalability, and guarantees predictable response latency, especially in presence of unreliable/unstable connectivity. Moreover, it may also improve energy efficiency and privacy, by avoiding the highly-consuming wireless transmission of large amounts of raw (and possibly sensitive) data to the cloud~\cite{edge_computing_2016}. However, running DL inference at the edge implies deploying models on cost-constrained devices such as microcontrollers (MCUs), with extremely limited memory and low operating frequencies.

In this context, TCNs are especially interesting due to their hardware-friendly properties. Nonetheless, their deployment at the edge still requires a careful tuning of all hyper-parameters, in order to meet the accuracy requirements of the target application at the minimum cost in terms of number of parameters or operations. For standard CNNs, popular in computer vision, this tuning is increasingly performed with automatic tools, in a process known as Neural Architecture Search (NAS). Recent years have seen the appearance of a plethora of different NAS approaches, based on as many different search strategies, some specifically targeting edge devices~\cite{nas_reinforcement_2016, fbnetv2_2020,morphnet_2018,mnasnet_2019,proxylessnas_2018}.

While TCNs have most of their hyper-parameters in common with standard CNNs (e.g. the filter sizes, the channels in each layer, etc.)
%
%
there is one fundamental peculiarity of these networks that requires an orthogonal approach, i.e., \textit{time dilation}. As explained in Sec.~\ref{subsec:tcn}, dilation allows enlarging the receptive field of a TCN on the time axis without increasing the number of parameters. 
Tuning dilation parameters, therefore, offers a clear pathway to make TCN topologies smaller and more hardware-friendly.

In this work, we propose Pruning In Time (PIT), a novel and light-weight automated method to simultaneously optimize the dilation of all layers in a TCN, producing Pareto-optimal solutions in terms of accuracy and complexity within a \textit{single} training run. This is achieved modeling the problem as a \textit{structured weight pruning along the time axis}: starting from maximally-sized filters with no dilation, we concurrently train the model weights and increase the dilation by pruning regularly-spaced weights slices on the time dimension. To the best of our knowledge, PIT is the first architecture optimizer that seamlessly targets dilation as the main optimization knob; state-of-the-art tools (e.g.~\cite{proxylessnas_2018}) require explicitly listing all possible parameter combinations for every layer. With experiments on two different time-series processing tasks, we show that our method can find dilation factors that reduce the model size and inference latency on the GAP8 System-on-Chip (SOC)~\cite{flamand2018gap} by up to 7.4$\times$ and 3$\times$ respectively, with no accuracy drop compared to a network without dilation. Moreover, it can produce a rich set of Pareto-optimal TCNs starting from a single seed, including solutions that outperform hand-designed models, reducing the number of parameters by up to 54\% without accuracy drop.

\section{Background and Related Works} 
\subsection{Temporal Convolutional Networks} \label{subsec:tcn}
TCNs share with 1-dimensional (1D) CNNs the main building blocks, i.e. convolutional, pooling and linear layers.
However, to improve the processing of time series, TCN convolutions include two new properties. 
\emph{Causality} forces the padding to be applied only on left side of the sequence; therefore, the outputs $\mathbf{y}_{t}$ are only functions of inputs $\mathbf{x}_{\tilde{t}}$ with $\tilde{t} \leq t$, so that output samples are not obtained taking information from the future.
\emph{Dilation} increases the receptive field of 1D convolutions without increasing the filters sizes, by applying a fixed step $d$ between the input samples processed by each filter.
Therefore, a 1D-convolution in a TCN is expressed by the following function:
\begin{equation}\label{eq:1d_conv}
\mathbf{y}_t^m = \text{Conv}\,(\mathbf{x}) = \sum_{i=0}^{K-1} \sum_{l=0}^{C_{in}-1} \mathbf{x}_{t-d\,i}^l \cdot \mathbf{W}_i^{l,m}
\end{equation}
where $t$ is the time index, $\mathbf{W}$ the filter weights, $C_{in}$ the number of input channels, $m$ the output channel, and $K$ the filter size.

\subsection{Neural Architecture Search}
Neural Architecture Search (NAS) is increasingly used to automatically design deep neural networks, avoiding the hand tuning of hyper-parameters, e.g. layer channels, filter dimensions, etc.
%
%
NAS automatically explores a large design space of possible hyper-parameters settings, co-optimizing the accuracy of the resulting network with its cost, measured as
the memory footprint~\cite{morphnet_2018}, the number of Floating Point Operations (FLOPs) or even the latency~\cite{proxylessnas_2018}. NAS outputs are sets of Pareto-optimal architectures in the complexity-accuracy space, from which users can select the best solution given their problem constraints.
Due to this capability, the exploration of NAS solutions tuned for extreme edge devices has recently started getting traction~\cite{mcunet_2020,proxylessnas_2018,onceforall_2020}.

We can categorize NAS methods in three main groups.
Early approaches used reinforcement learning or evolutionary algorithms to improve the structure of the network at the cost of thousands of training runs (and GPU hours), thus being often prohibitive for real-world tasks~\cite{nas_reinforcement_2016}.
More recent research has focused on \emph{differentiable} NAS solutions (DNAS)~\cite{darts_2019}, that model the problem as the optimization of a so-called \textit{supernet}, which includes different alternative implementations of each layer.
%
%
These methods jointly train a set of binary variables that determines a selection of layer implementations from the supernet (called a \textit{path}) and the weights of that path.
%
%
Although more efficient than the previous group, memory occupation and training time remain critical for these algorithms, since the supernet includes all possible combinations of layers implementations.
%
ProxylessNAS~\cite{proxylessnas_2018} tackles the memory problem by training a single path of the supernet per batch, thus limiting the memory occupation of the whole model, but not exploring the full solution space at the same time.

Lastly, an emerging set of DNAS methods is based on so-called DMaskingNAS~\cite{fbnetv2_2020, morphnet_2018}, which aims at reducing the training time by limiting the search space to a single \emph{seed} network.
DMaskingNAS approaches add trainable parameters (``masks'') to the seed network to tweak architectural features such as the number of channels or the filter dimensions.
%
Specifically, a vector of $\gamma_{i}^{(l)}$ is defined per each layer $l$, and combined with the normal weights of the model, in such a way that setting each of the $\gamma_{i}^{(l)}$ to zero has an effect on the architecture (e.g. eliminating a channel or reducing the filter size). Then, the non-zero $\gamma_{i}^{(l}$ are minimized during training with an approach similar to \textit{weight pruning}, thus reducing the network size.
%
%
For instance, MorphNet~\cite{morphnet_2018} uses the already present multiplicative terms in batch normalization (BN) layers as $\gamma_i^{(l)}$, to zero-out entire channels.
%
%
While the search space of DMaskingNAS algorithms is slightly more restricted than DNAS, it is still large enough to produce high-quality solutions; and at the same time, it can be explored at similar cost to a \textit{single} network training.

In our work, we propose the first DMaskingNAS solution tuned specifically at exploiting the main architectural feature of TCNs -- namely, \textit{dilation} -- to find Pareto-optimal alternatives in terms of accuracy/model size, to ease deployment on memory-starved IoT devices.


%
%

\section{Pruning In Time}

Despite the large number of NAS methods recently proposed in literature, most of them have focused on optimizing hyper-parameters that are relevant for standard CNNs, such as the number of channels in each layer and the kernel sizes~\cite{morphnet_2018,fbnetv2_2020,proxylessnas_2018}, with particular focus on 2D CNNs for computer vision. To the best of our knowledge, no existing NAS has explicitly targeted the automatic optimization of dilation factors in 1D CNNs ($d$ in Eq.~\ref{eq:1d_conv}). Nonetheless, dilation is fundamental for the accuracy of these networks, as detailed in~\cite{tcn_original_2018}, since it controls the relation between their receptive field in time (i.e., the range of samples covered by each convolution step) and their filter sizes (which determine the model complexity). Setting dilation factors by hand requires an empirical and time consuming trial and error process.

\begin{figure}[t!]
  \centering
\includegraphics[width=.95\columnwidth]{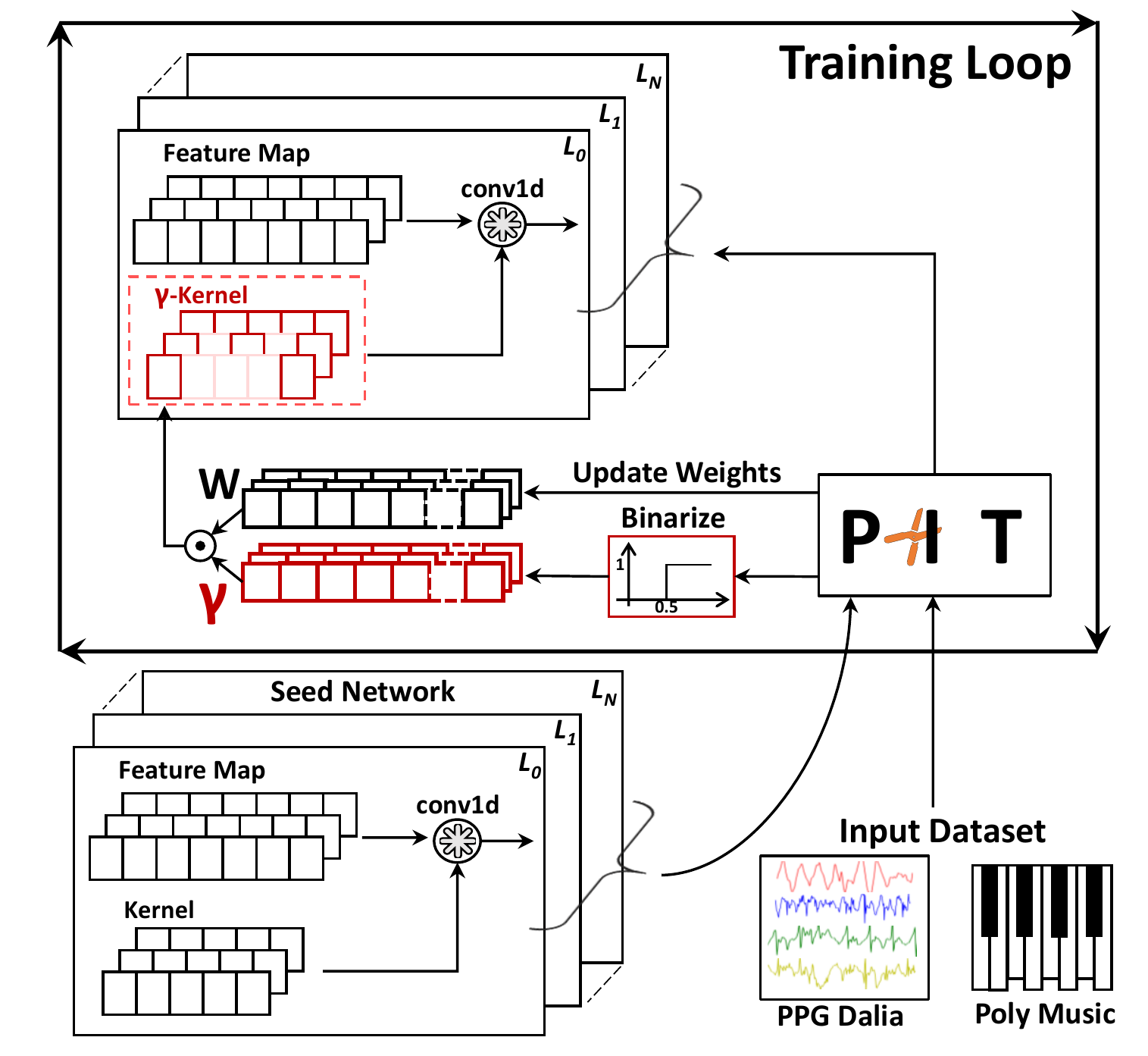}
  \caption{Training flow of the proposed Pruning In Time architecture optimizer.}
  \label{fig:flow}
\end{figure}

In this paper, we make up for this lack by proposing \textit{Pruning In Time} (PIT), a light-weight and efficient DMaskingNAS method that learns the optimal set of dilation factors for a TCN given as seed network. 
The functionality of PIT is shown in Fig.~\ref{fig:flow}: our algorithm starts from a seed network with maximally-sized filters and dilation $d=1$ in all layers, and concurrently learns the TCN weights and the optimal dilations for each layer, based on a cost metric.
Specifically, in this work we consider model size as the target metric, but the method is easily extendable to other types of optimizations (e.g., FLOPs reduction). The optimization of dilation factors is obtained by modeling the problem as a \textit{weight pruning in time}, as explained in the following, hence the name of our approach.

\begin{figure}[t!]
  \centering
    \includegraphics[width=.99\columnwidth]{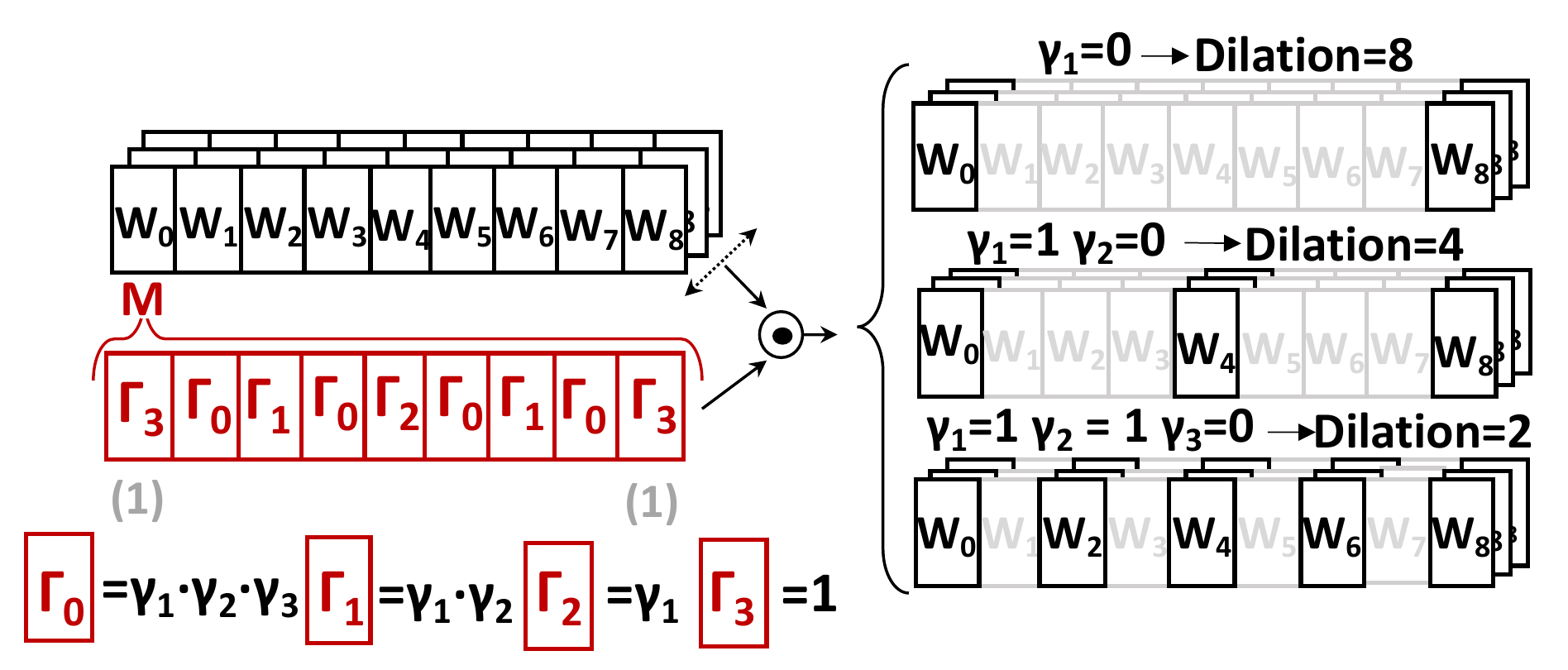}
  \caption{Combination of $\bm{\gamma}$ elements with each other and with convolution filter weights to form different dilation patterns. Example for $\mathit{rf}_\mathrm{max}=9$.}
  \label{fig:dil_prun}
   \vspace{-0.6cm}
\end{figure}
\subsection{Making Dilation Differentiable}
%
%
In order to build a DNAS algorithm for dilated convolutions, the primary issue to be solved is how to embed the different dilation factors in a (differentiable) component of the loss function, so that they can be optimized during training. 
Similarly to other DMaskingNAS techniques, our approach is built upon additional trainable parameters, which we call $ \bm{\gamma} $. 

Each temporal convolution in a TCN is characterized by a certain kernel size $ k $ and a certain dilation factor $ d $. Together, these two parameters define the receptive field $ \mathit{rf} $ of the layer: $ \mathit{rf} = (k - 1) \cdot d $.
Notice that, in this formulation, the dilation is \textit{regular}, i.e., the number of skipped time-steps is constant in each convolutional layer.
PIT limits the search space to such solutions, which are the only ones supported by current-generation inference libraries for MCUs~\cite{CubeAi,NNtool}, and generally enable better low-level optimizations and more regular memory access patterns.
In particular, we focus on dilation factors $d$ that are expressed as powers of 2.
These maintain a high degree of freedom in enlarging or reducing the receptive field, while restricting the solution space and enabling the simple formalization that we treat in the following.

For every temporal convolution, PIT starts by defining a vector of \textit{binary parameters} $ \bm{\gamma} $, containing $  L = \lfloor{\log_{2}(\mathit{rf}_\mathrm{max} - 1)}\rfloor + 1$ elements, where $\mathit{rf}_\mathrm{max}$ is the maximum supported $\mathit{rf}$. In particular, $\gamma_0$ is constant and always equal to 1, and is added just to simplify the mathematical notation.
The remaining $\gamma_{1:L-1}$, instead, are the main knobs that control dilation.
%
%
We binarize $\bm{\gamma}$ following a BinaryConnect-like approach~\cite{courbariaux2016binarized}. In forward-propagation, we use a Heaviside step function and a threshold $\delta$, fixed to 0.5 in our experiments:
\begin{equation}
\mathcal{H}(\hat{\gamma_i} - \delta) = 
\begin{cases}
  1,     & \text{\textit{for} } \hat{\gamma_i} \ge \delta  \\    
  0, & \text{\textit{for} } \hat{\gamma_i} < \delta
\end{cases}
\end{equation}
where $\hat{\gamma_i}$ is the floating point version of $\gamma_i$. In backward passes, because the derivative of this function is zero almost everywhere, its gradient is treated with a straight-through estimator~\cite{courbariaux2016binarized}, i.e., the step is replaced with an identity function for the purpose of propagating gradients.

To restrict the search space to regular dilation patterns, the trainable parameters in $\bm{\gamma} $ are combined together in the way depicted in the red part of Fig.~\ref{fig:dil_prun}.
We start by multiplying together the elements in $\bm\gamma$ to form a new set of $\bm\Gamma$, which will be the actual values used to perform the masking:
%
%
\begin{equation}\label{eq:gamma_mult}
\bm{\Gamma_i} = \prod_{k=0}^{L-1-i} \bm{\gamma_{k}}
\end{equation}
%
%
%
Notice that, in Fig.~\ref{fig:dil_prun}, $\gamma_0$ has been directly replaced with 1.
Intuitively, the parameters $\bm\Gamma$ have to act as on/off selectors that enable or disable entire time slices of the convolution filter, encoding regular patterns of power-of-two dilation.

As shown in the figure, $d=1$ is achieved only when all trainable parameters are 1, i.e., when $\Gamma_0 = \gamma_0 \cdot \gamma_1 \cdots \gamma_{L-1} = 1$.
To encode patterns with larger dilation, every time we double $d$ we remove the condition on one $\gamma_i$, so for $d=2$ we use $\Gamma_1 = \gamma_0 \cdot \gamma_1 \cdots \gamma_{L-2} = 1$, etc., up to the highest supported dilation ($d=2^{L-1}$), which is obtained with the (always true) condition $\Gamma_{L-1} = \gamma_0 = 1$.

To obtain this behavior, we further combine the elements of $\bm{\Gamma}$ in a \textit{mask vector} $\bm{M}$, of length $\mathit{rf}_\mathrm{max}$, as shown in Fig.~\ref{fig:dil_prun}. Each element of $\bm{M}$ is then multiplied with \textit{all filter weights} relative to the corresponding time-step (and all channels) to perform the actual masking.
%
%
The right side of Fig.~\ref{fig:dil_prun} shows how the various dilation alternatives are mapped to the trainable $\bm\gamma$ through the masking procedure described above, for the case of $\mathit{rf}_\mathrm{max}=9$ ($L=4$).

\begin{figure}[t!]
  \centering
\includegraphics[width=\columnwidth]{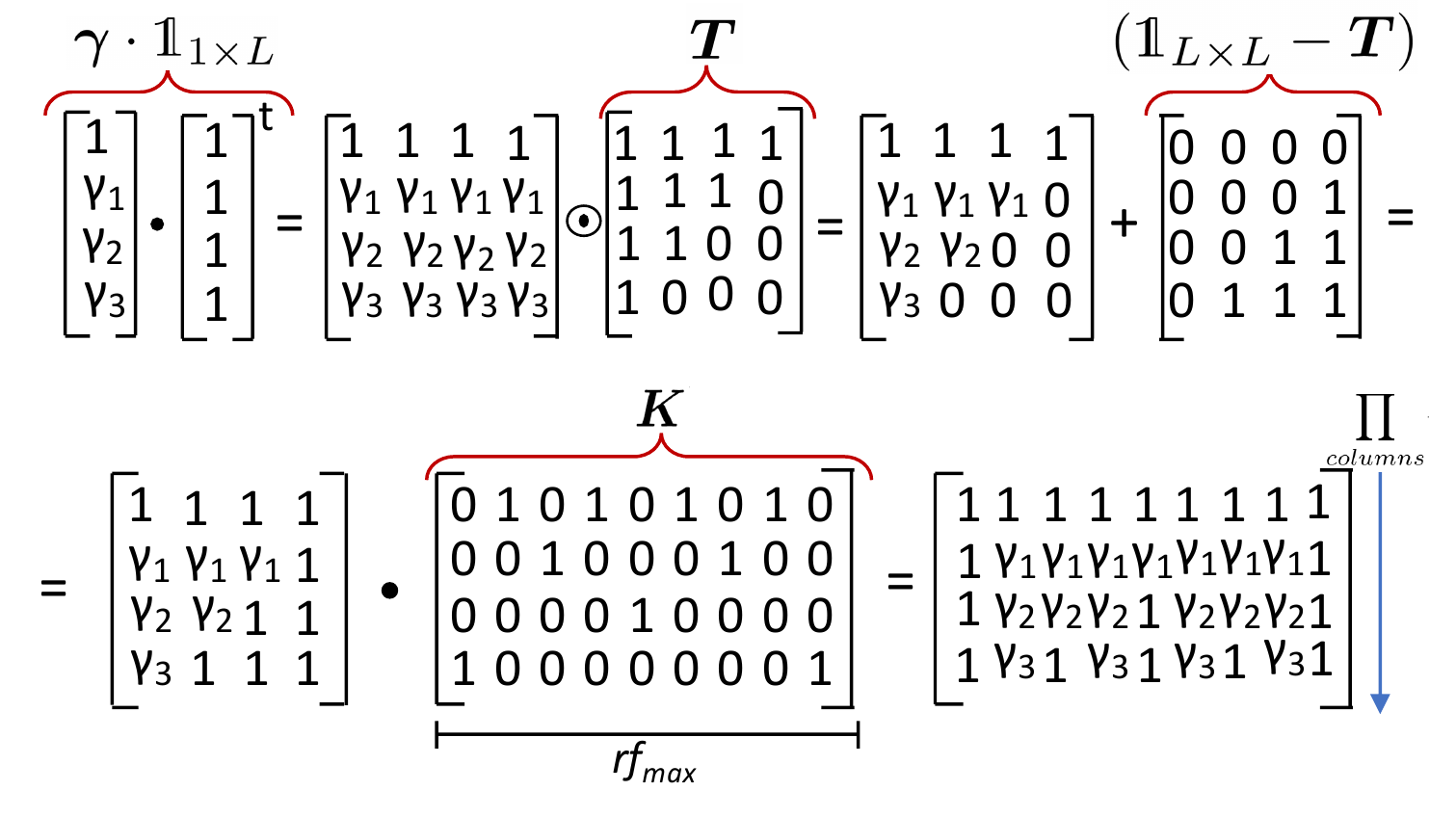}
   \vspace{-0.6cm}
  \caption{Example of generation of the $\bm{M}$ mask vector with differentiable tensor operations, for the same layer size as Fig.~\ref{fig:dil_prun}.}\label{fig:M_calc}
   \vspace{-0.6cm}
\end{figure}

In order to train $\bm{\gamma}$ with a DNAS method, the constructive description of the mask vector $\bf{M}$ given above must be expressed in a differentiable form.
To do so, we use the following tensor transformation:
\begin{equation}\label{eq:gamma_transf}
\bm{M} = \prod_{columns} \{ [(\bm{\gamma}  \cdot \mathds{1}_{1 \times L}) \odot \bm{T} +
(\mathds{1}_{L \times L} - \bm{T})] \cdot \bm{K} \}
\end{equation}
where $\prod_{columns}$ indicates the product of all elements in each column of the final matrix, $\mathds{1}_{i \times j} $ is a matrix of 1s of size $ i \times j$ and $ \odot $ is the Hadamard product. $\bm{T}$ and $\bm{K}$ are two constant matrices of 0s and 1s. An example of these two matrices and of the entire transformation is shown in Fig.~\ref{fig:M_calc}, for the same $\mathit{rf}_\mathrm{max}$ and $L$ used in Fig.~\ref{fig:dil_prun}. Again, $\gamma_0$ has been directly replaced with 1 for clarity. Notice that $\bm{T}$ is just an upper triangular matrix with inverted columns, whereas $\bm{K}$ can be generated procedurally for any value of $\mathit{rf}_\mathrm{max}$, by repeating a pattern of 0s and 1s (procedure not shown for sake of space).



%

Finally, the complete layer equation for dilated convolution in PIT becomes:
\begin{equation} \label{eq:gamma_conv}
\mathbf{y}_t^m = \sum_{i=0}^{\mathit{rf}_\mathrm{max}-1} \sum_{l=0}^{C_{in}-1} \mathbf{x}_{t-i}^l \cdot (\bm{M_i} \odot \mathbf{W}_i^{l,m})
\end{equation}
where all symbols have been defined in previous Eq.~\ref{eq:1d_conv}-\ref{eq:gamma_transf}.

\subsection{Dilation Regularizer}\label{sec:loss}

Having modified all convolutional layers with $ \bm{\gamma} $ vectors, PIT performs a standard training, where the loss function is augmented with a Lasso regularization term, to promote the sparsification (pruning) of the $ \bm{\gamma}_{i} $, thus enabling the exploration of architectures with $d>1$.
%

%
The regularization term can be customized to direct the search towards a target cost metric. In this work, we consider the network size as our proxy for cost. Therefore, we use the following regularizer:
\begin{equation} \label{eq:lasso_reg}
\mathcal{\bm{L}}_{R}^{size}(\bm{\gamma}) = \lambda \sum_{l=1}^{layers}C_{in}^{(l)} \cdot C_{out}^{(l)} \sum_{i=1}^{L-1} \mathrm{round}\left( \frac{\mathit{rf}_\mathrm{max} - 1}{2^{L-i}} \right) \lvert \bm{\hat{\gamma}}_{i}^{(l)} \rvert  
\end{equation}
where $ \lambda $ controls the strength of the regularization and superscript $(l)$ refers to the $l$-th layer. $C_{in}^{(l)}$ and $C_{out}^{(l)}$ are the number of input/output channels in the $l$-th layer, both of which influence the layer size in a way which is linearly proportional to the amount of non-pruned time-slices in the filter matrix.
Finally,  $\mathrm{round} \left( \frac{\mathit{rf}_\mathrm{max} - 1}{2^{L-i}} \right)$ is the number of alive (i.e. non-masked) filter time-slices added by each non-zero $\gamma_{i}^{(l)}$ (see Fig.~\ref{fig:dil_prun}).
%
%

Globally, the loss function optimized in the pruning phase by PIT is the following:
\begin{equation}
    \mathcal{\bm{L}}_{PIT} (\bm{W}, \bm{\gamma})  = 
    \mathcal{\bm{L}}^{perf} (\bm{W})  + \mathcal{\bm{L}}_{R}^{size} (\bm{\gamma}) 
\end{equation}
where $\mathcal{\bm{L}}^{perf} (\bm{W})$ is the loss term related to the TCN's performance, e.g., the classification accuracy.

\subsection{Training Procedure}

Algorithm~\ref{alg:PIT} summarizes the training procedure implemented in PIT. As shown, the process is composed of three separate phases. Initially, we perform a warmup lasting $\rm Steps_{wu}$ training steps. In this phase, all elements in $\bm{\gamma}$ vectors are initialized to 1. Therefore, we only train the weights of the seed network, with maximally sized filters and $d=1$ in all convolution layers, and considering only the network performance as objective.
Next, the main pruning loop is initiated. Here, we concurrently update the weights and the $\bm{\gamma}$ vectors with the regularized loss. Once the pruning phase reached convergence (i.e., loss not improving on the validation set), we enter the third phase, where all $\bm{\gamma}$ are frozen to their latest binarized values, and the resulting network with dilation is fine-tuned, again considering only performance in the loss.
We found that both warmup and fine-tuning significantly improve the final accuracy of the pruned networks.

 \begin{algorithm}
 \begin{algorithmic}[1]
 \caption{\label{alg:PIT} PIT - Pruning in Time}
 \For{$i \gets 1, \dots, \rm Steps_{wu}$} {\color{darkspringgreen}\#warmup loop}
     \State Update $\bm{W}$ based on $\nabla_{\bm{W}} \mathcal{L}_{perf}(\bm{W})$
 \EndFor
 \While{not converged} {\color{darkspringgreen}\#pruning loop}
     \State Update $\bm{W}$ and $\bm{\gamma}$ based on $\nabla_{\bm{W}, \bm{\gamma}} \mathcal{L}_{PIT}(\bm{W}, \bm{\gamma})$
 \EndWhile
 \For{$i \gets 1, \dots, \rm Steps_{ft}$} {\color{darkspringgreen}\#fine-tuning loop}
     \State Update $\bm{W}$ based on $\nabla_{\bm{W}} \mathcal{L}_{perf}(\bm{W})$
 \EndFor
 \end{algorithmic}
 \end{algorithm}

The procedure of Algorithm~\ref{alg:PIT} can be repeated with different regularizer strenghts ($\lambda$ in Eq.~\ref{eq:lasso_reg}) to explore the performance versus cost design space. Similarly, also the number of warmup steps has an impact on the same trade-off. Indeed, as explained in~\cite{proxylessnas_2018}, a shorter warmup, yielding a less accurate network at the beginning of pruning, tends to favor model simplifications, as their impact on accuracy is less at the beginning of the pruning phase.

Importantly, PIT can be easily integrated with other DMaskingNAS techniques that affect different hyper-parameters, e.g., \cite{morphnet_2018} to tune the number of channels in each layer, simply by adding further regularization terms and masking parameters, to perform a wider exploration.

\section{Experimental Results}
\subsection{Setup}
\textbf{Datasets \& architectures.}
We benchmark PIT using two different datasets and 1D-CNN seed networks.
The first is Notthingham, a polyphonic music dataset extracted from 1200 American and British folk tunes~\cite{tcn_original_2018}.
Each input is a sequence of samples, each represented by a 88-bit binary code, corresponding to the 88 keys of a piano.
As a starting point to build the seed model for this dataset, we use the TCN presented in~\cite{tcn_original_2018}, here refered as ResTCN. 
In detail, to automatically search for the best dilation parameters, we start from ResTCN with identical receptive fields as the one of \cite{tcn_original_2018}, but setting $d = 1$ in each layer. 
%
The second dataset is PPGDalia~\cite{reiss2019deep}, the largest publicly available dataset for PPG-based heart rate estimation. 
It includes measurements from a PPG-sensor and a 3D-accelerometer, together with golden HR values for 15 subject and a total of 37.5 hours of recording.
In this case we use TEMPONet, first introduced in \cite{temponet_2019}, as our seed architecture, since this network achieved excellent results on similar bio-signal processing tasks. Again, the seed is obtained setting $d=1$ in all layers while maintaining the receptive fields.
The PIT code and the results for all the other benchmarks introduced in~\cite{tcn_original_2018}
are released open source at \texttt{https://github.com/matteorisso/PIT} and are not reported here for sake of space.

\textbf{Deployment.} To measure latency and energy gains obtained by PIT, we deploy models produced by our tool on the GreenWaves Technologies' GAP8 System-on-Chip (SoC)~\cite{flamand2018gap}, a parallel ultra-low-power platform that features one I/O core and an 8-core cluster with a RISC-V Instruction Set Architecture extension for enhanced digital signal processing.
The SoC includes a two level memory hierarchy: a single-cycle clock latency 64 kB L1 scratchpad, and a 512 kB L2 memory. Additionally, an off-chip L3 memory can be plugged to expand the storage capability.  
GAP8 also includes two general-purpose Direct Memory Access (DMA) controllers to move data between memories, reducing the memory access bottleneck.
%

We deploy int8-quantized models using GreenWaves' proprietary neural network deployment flow, called NN-Tool, which supports dilated 1D convolutions.
We set the I/O core frequency and the 8-core cluster frequency to 100 MHz.

\subsection{Design Space Exploration}
\begin{figure}[t!]
  \centering
\includegraphics[width=.95\columnwidth]{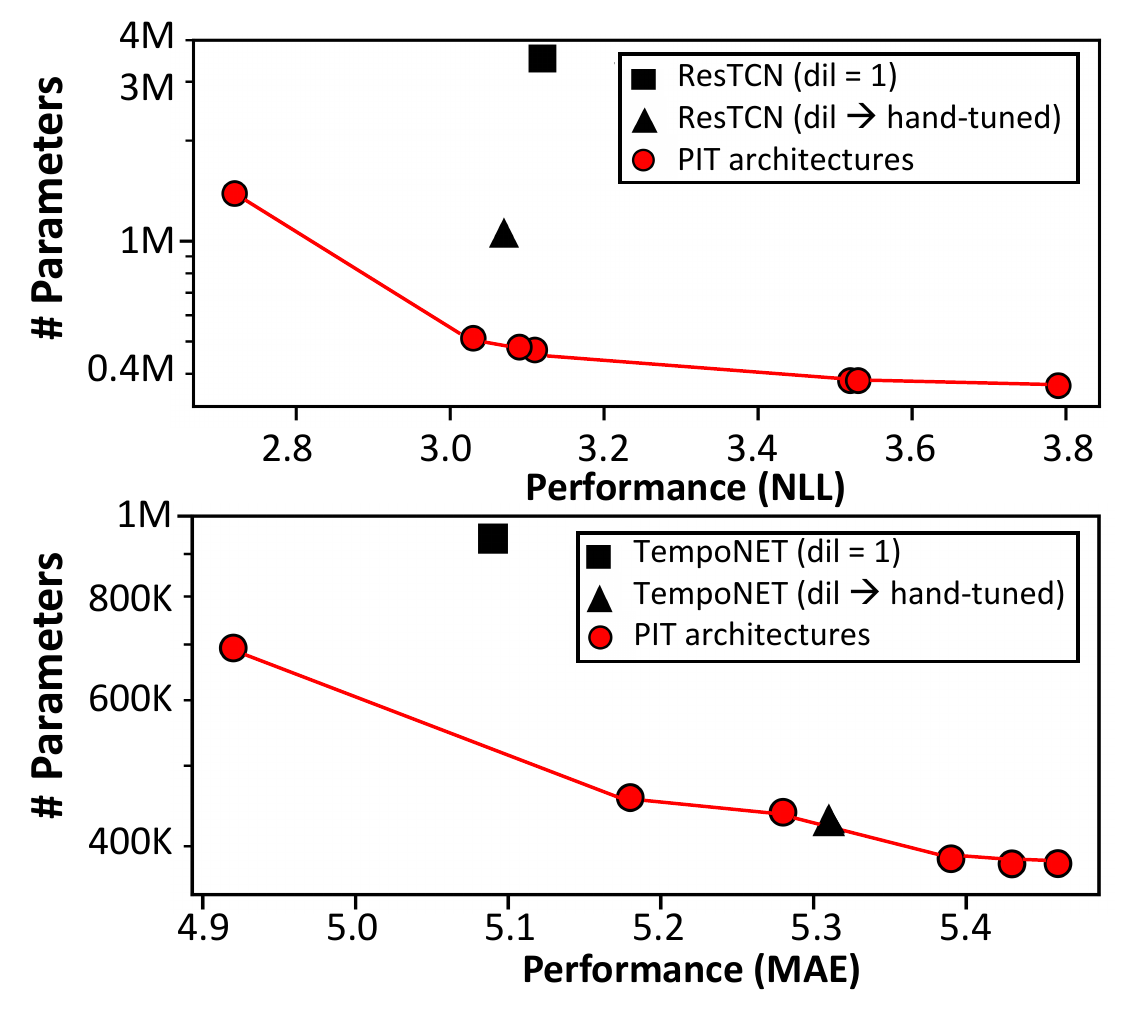}
  \vspace{-0.5cm}
  \caption{PIT Pareto frontiers obtained starting from two seed architectures on the Nottingham dataset (top) and on PPGDalia (bottom). Each plot also reports the seed model (square) and the original hand-engineered TCN (triangle).}
  \label{fig:PIT_res}
  \vspace{-0.5cm}
\end{figure}

\begin{table}
\centering
\caption{Dilations obtained for the different convolutional layers of ResTCN and TEMPONet in different PIT outputs.}
\label{tab:PIT_dilations}
\begin{tabular}{|l|l|}
\hline 
network                 & PIT dilations               \\\hline
ResTCN dil=hand-tuned      & (1, 1, 2, 2, 4, 4, 8, 8)         \\
PIT ResTCN small              & (4, 4, 8, 8, 16, 16 32, 32) \\
PIT ResTCN medium              & (4, 1, 4, 8, 16, 16, 32, 32)         \\
PIT ResTCN large              & (1, 4, 8, 8, 16, 16, 8, 1)           \\ \hline \hline
TEMPONet dil=hand-tuned & (2, 2, 1, 4, 4, 8, 8)               \\
PIT TEMPONet small         & (2, 4, 4, 8, 8, 16, 16)             \\
PIT TEMPONet medium         & (1, 2, 4, 2, 1, 8, 16)              \\
PIT TEMPONet large         & (1, 1, 1, 1, 1, 1, 16)             \\\hline 
\end{tabular}
\vspace{-0.4cm}
\end{table}
Starting from the two seed architectures detailed above, we perform a complete design space exploration, by tweaking the $\lambda$ regularization-strength of PIT and the warmup duration (Steps$_{wu}$). 
Results are summarized in Fig.~\ref{fig:PIT_res}.

Considering all possible power-of-two dilations achievable given the seed network's receptive fields, PIT operates in a search space of $ {\sim}10^{5}$ different solutions for the ResTCN, spanning from NNs with 400k to 3M parameters.
For TEMPONet, the search includes $ {\sim}10^{4} $ alternatives, ranging from 400k to 900k parameters.
The dilations obtained by PIT for three different architectures, namely the largest (large), the smallest (small) and the closest in size to the original hand-designed ResTCN/TEMPONet (medium) are reported in Table~\ref{tab:PIT_dilations}.
For both benchmarks, PIT finds new Pareto-optimal architectures that improve both the accuracy and the network size simultaneously, compared to the seed baselines without dilation (black squares in the figure).
In detail, we find a ResTCN-variant with $ 2.54\times $ less parameters and an accuracy improvement of 13\% (from 3.12 to 2.72 Negative Log-Likelihood - NLL loss) with respect to the seed; concerning TEMPONet, the top-performing architecture shows a compression of $ 1.35\times $ with a Mean Absolute Error (MAE) reduction of 0.16.
PIT also finds a new ResTCN-based architecture that dominates the \textit{hand-tuned} original network presented in \cite{tcn_original_2018} in the Pareto sense, resulting in a 54\% reduction of the parameters with almost identical performance. While the same is not true for TEMPONet, the hand-engineered network sits on the Pareto frontier in this case, showing the good quality of the architectures identified by PIT.
%
%
\subsection{Comparison with state-of-the-art}
\begin{table}
\centering
\caption{Comparison between PIT and ProxylessNAS, with TEMPONet as seed architecture and PPGDalia as dataset.}
\label{tab:comparison}
\begin{tabular}{l|l|l|l|l|}
\cline{2-5}
                             & \multicolumn{2}{c|}{ProxylessNAS} & \multicolumn{2}{c|}{Pruning in Time} \\ \cline{2-5} 
                             & \# weights            & Perf. [MAE]            & \# weights              & Perf. [MAE]              \\ \hline
\multicolumn{1}{|l|}{small}  &        381k          &      5.43          & 381k              & 5.43             \\ \hline
\multicolumn{1}{|l|}{medium} &         517k         &        5.21        & 440k              & 5.28             \\ \hline
\multicolumn{1}{|l|}{large}  &          731k        &       5.15         & 694k              & 4.92             \\ \hline
\end{tabular}
\end{table}
\begin{figure}[t!]
  \centering
\includegraphics[width=.95\columnwidth]{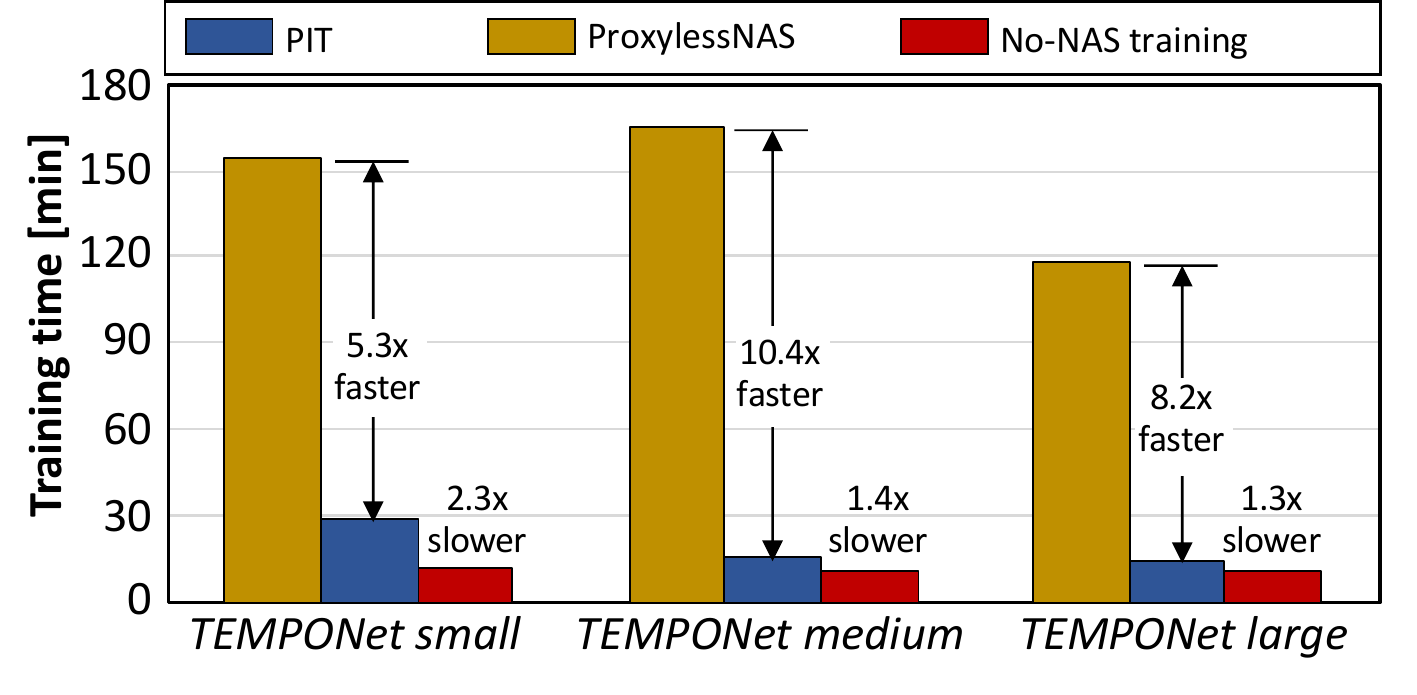}
  \vspace{-0.5cm}
  \caption{Comparison of training time for ProxylessNAS, PIT, and a single NN training using a seed TEMPONet network and the PPGDalia dataset.}
  \label{fig:proxy_pit_time}
  \vspace{-0.5cm}
\end{figure}
Table~\ref{tab:comparison} compares solutions found by PIT with the ones of the state-of-the-art ProxylessNAS \cite{proxylessnas_2018}, a DNAS algorithm based on the supernet idea, that can be adapted to search over different dilation factors in a 1D-CNN by manually including all layer variants in the supernet.
Specifically, for each layer, ProxylessNAS chooses between multiple alternatives, that we manually specified by increasing $d$ and keeping $C_{in}$ and $C_{out}$ constant, so to match exactly the search space explored by PIT. 
For sake of space, the comparison is reported only on TEMPONet as seed architecture with PPGDalia as dataset.
In Table~\ref{tab:comparison}, we report three different architectures found by each algorithm (small, medium and large, selected with the same rationale of Table~\ref{tab:PIT_dilations}).
Both algorithms converge to same  \textit{small} network, while solutions are similar for the other two cases.
Noteworthy, in the \textit{large} case, PIT finds an architecture that is both smaller (694k vs. 731k parameters) and more accurate (4.92 vs 5.15 MAE) than ProxylessNAS.

In Fig.~\ref{fig:proxy_pit_time}, we compare the training time required by PIT and ProxylessNAS to find the three networks.
All the experiments have been performed with the same hardware set-up, namely a single \textit{NVIDIA-GTX 1080Ti} GPU, and the same training algorithm parameters, including batch size = 128 and an early-stop patience of 50 epochs.
PIT reduces the search time compared to ProxylessNAS by up to 10.4$\times$, being only 1.3$\times$-2.3$\times$ slower than the training of a single hand-designed TEMPONet architecture.
This is mainly due to the concurrent training of \textit{all weights} and masking parameters performed by PIT.
%
%
%
In contrast, ProxylessNAS trains only one path of the supernet (and the corresponding weights) in each iteration of the training loop, causing a significant time overhead.

\subsection{Deployment on GAP8}
\begin{table}
\centering
\caption{PIT solutions compared to original seed networks without dilation and with hand-tuned dilation deployed on GAP8 SoC.}
\label{tab:deployment}
\begin{tabular}{l|l|l|l|l|}
\cline{2-5}
                                              & \# weights & loss & latency & energy  \\ \hline
\multicolumn{1}{|l|}{ResTCN dil=1}               & 3.53M      & 3.12 NLL               & 1002 ms             & 262.7 mJ           \\
\multicolumn{1}{|l|}{ResTCN dil=h.-t.}      & 1.05M      & 3.07 NLL               & 500 ms              & 131 mJ             \\
\multicolumn{1}{|l|}{PIT ResTCN s.}              & \textbf{0.37M}      & 3.79 NLL              & 336.7 ms            & 88.2 mJ            \\
\multicolumn{1}{|l|}{PIT ResTCN m.}              & 0.48M      & 3.09 NLL               & \textbf{335.9 ms}            & \textbf{87.9 mJ}            \\
\multicolumn{1}{|l|}{PIT ResTCN l.}              & 1.39M      & \textbf{2.72 NLL}               & 539.2 ms            & 141.3 mJ           \\ \hline \hline
\multicolumn{1}{|l|}{TEMPONet dil=1}          & 939K       & 5.08 MAE               & 112.6 ms            & 29.5 mJ            \\
\multicolumn{1}{|l|}{TEMPONet dil=h.-t.} & 423K       & 5.31 MAE               & 58.8 ms             & 15.4 mJ            \\
\multicolumn{1}{|l|}{PIT TEMPONet s.}         & \textbf{381K}       & 5.43 MAE               & \textbf{54.8 ms}             & \textbf{14.4 mJ}            \\
\multicolumn{1}{|l|}{PIT TEMPONet m.}         & 440K       & 5.28 MAE               & 59.8 ms             & 15.7 mJ            \\
\multicolumn{1}{|l|}{PIT TEMPONet l.}         & 694K       & \textbf{4.92 MAE}               & 86.3 ms             & 22.6 mJ           \\ \hline
\end{tabular}
\vspace{-0.5cm}
\end{table}
We deployed the original hand-engineered TCNs, the seed networks with unitary dilation, and the three different PIT outputs for each dataset reported in Table~\ref{tab:PIT_dilations}, on the 8-cores cluster of GAP8.
Table~\ref{tab:deployment} reports the results in terms of network size, loss (either NLL or MAE for the two datasets), latency and energy.
%
%
%

Deploying the ``medium'' PIT ResTCN, we obtain a loss equivalent to that of the hand-tuned network presented in \cite{tcn_original_2018} with 2.2$\times$ less parameters, and 1.5$\times$ lower latency and energy.
For TEMPONet, we obtain a medium network which is equivalent to the hand-engineered one in every metric. 

Noteworthy, our small (medium) ResTCN networks have a 20\% higher (0.9\% lower) loss compared to the \textit{seed} network, but are 9.5$\times$ (7.4$\times$) smaller and 3.0$\times$ (3.0$\times$) faster.
On TEMPONet, the small (medium) models obtain a 6.9\%  (3.9\%) worse error compared to the seed, with 2.5$\times$ (2.1$\times$) less weights and 2.1$\times$ (1.9$\times$) lower latency and energy.
%
\section{Conclusions}
NAS methods are becoming essential tools for deep learning, enabling a fast exploration of the model architecture design space without the tedious trial-and-error approach that characterized the development of new neural networks in the past.
Therefore, it is fundamental to develop efficient NAS methods, which do not require extreme hardware resources and/or an enormous training times. At the same time, these tools must be capable of exploring all important hyper-parameters, which in case of a TCN include the dilation.
The proposed PIT is a light-weight NAS framework that goes in that direction, and successfully generates size-optimized and hardware-friendly TCNs.
PIT is able to generate improved versions of existing state-of-the-art architectures, with a compression of up to 54\% with negligible accuracy drop, enabling their efficient deployment on resource-constrained MCUs.

\footnotesize
\bibliographystyle{IEEEtran}


\end{document}